%% file: main_final.tex
\documentclass[10pt,twocolumn,letterpaper]{article}

\usepackage{iccv}
\usepackage[accsupp]{axessibility}
\iccvfinalcopy 
\input{config}

\def\iccvPaperID{8366} 

\ificcvfinal\fi

\begin{document}

\title{\vspace{-12pt}Spatio-temporal Self-Supervised Representation Learning for 3D Point Clouds}

\author{\vspace{3pt}Siyuan Huang$^{1,*}$, Yichen Xie$^{2,*}$, Song-Chun Zhu$^{3,4,5}$, Yixin Zhu$^{3,4}$\\
$^{1}$ University of California, Los Angeles $^{2}$ Shanghai Jiao Tong University \\\vspace{3pt} $^{3}$ Beijing Institute for General Artificial Intelligence
$^{4}$ Peking University $^{5}$ Tsinghua University\\
\url{https://siyuanhuang.com/STRL}\vspace{-15pt}\\
}

\maketitle

\blfootnote{* indicates equal contribution.}

\setstretch{0.98}

\begin{abstract}
To date, various 3D scene understanding tasks still lack practical and generalizable pre-trained models, primarily due to the intricate nature of 3D scene understanding tasks and their immense variations introduced by camera views, lighting, occlusions, \etc. In this paper, we tackle this challenge by introducing a \acf{strl} framework, capable of learning from unlabeled 3D point clouds in a self-supervised fashion. Inspired by how infants learn from visual data in the wild, we explore the rich spatio-temporal cues derived from the 3D data. Specifically, \ac{strl} takes two temporally-correlated frames from a 3D point cloud sequence as the input, transforms it with the spatial data augmentation, and learns the invariant representation self-supervisedly. To corroborate the efficacy of \ac{strl}, we conduct extensive experiments on three types (synthetic, indoor, and outdoor) of datasets. Experimental results demonstrate that, compared with supervised learning methods, the learned self-supervised representation facilitates various models to attain comparable or even better performances while capable of generalizing pre-trained models to downstream tasks, including 3D shape classification, 3D object detection, and 3D semantic segmentation. Moreover, the spatio-temporal contextual cues embedded in 3D point clouds significantly improve the learned representations.
\end{abstract}

\section{Introduction}

Point cloud is a quintessential 3D representation for visual analysis and scene understanding. It differs from alternative 3D representations (\eg, voxel, mesh) as it is ubiquitous: Entry-level depth sensors (even on cellphones) directly produce point clouds before triangulating into meshes or converting to voxels, making it mostly applicable to 3D scene understanding tasks such as 3D shape analysis~\cite{chang2015shapenet}, 3D object detection and segmentation~\cite{song2015sun,dai2017scannet}. Despite its omnipresence in 3D representation, however, annotating 3D point cloud data is proven to be much more difficult compared with labeling conventional 2D image data; this obstacle precludes its potentials in 3D visual tasks. As such, properly leveraging the colossal amount of \emph{unlabeled} 3D point cloud data is a \emph{sine qua non} for the success of \emph{large-scale} 3D visual analysis and scene understanding.

Meanwhile, self-supervised learning from unlabeled images~\cite{doersch2015unsupervised,oord2018representation,hjelm2018learning,he2020momentum,chen2020simple,grill2020bootstrap,chen2021exploring} and videos~\cite{sermanet2018time,zhuang2020unsupervised,knights2020temporally,qian2020spatiotemporal} becomes a nascent direction in representation learning with great potential in downstream tasks.

\begin{figure}[t!]
    \centering
    \includegraphics[width=\linewidth]{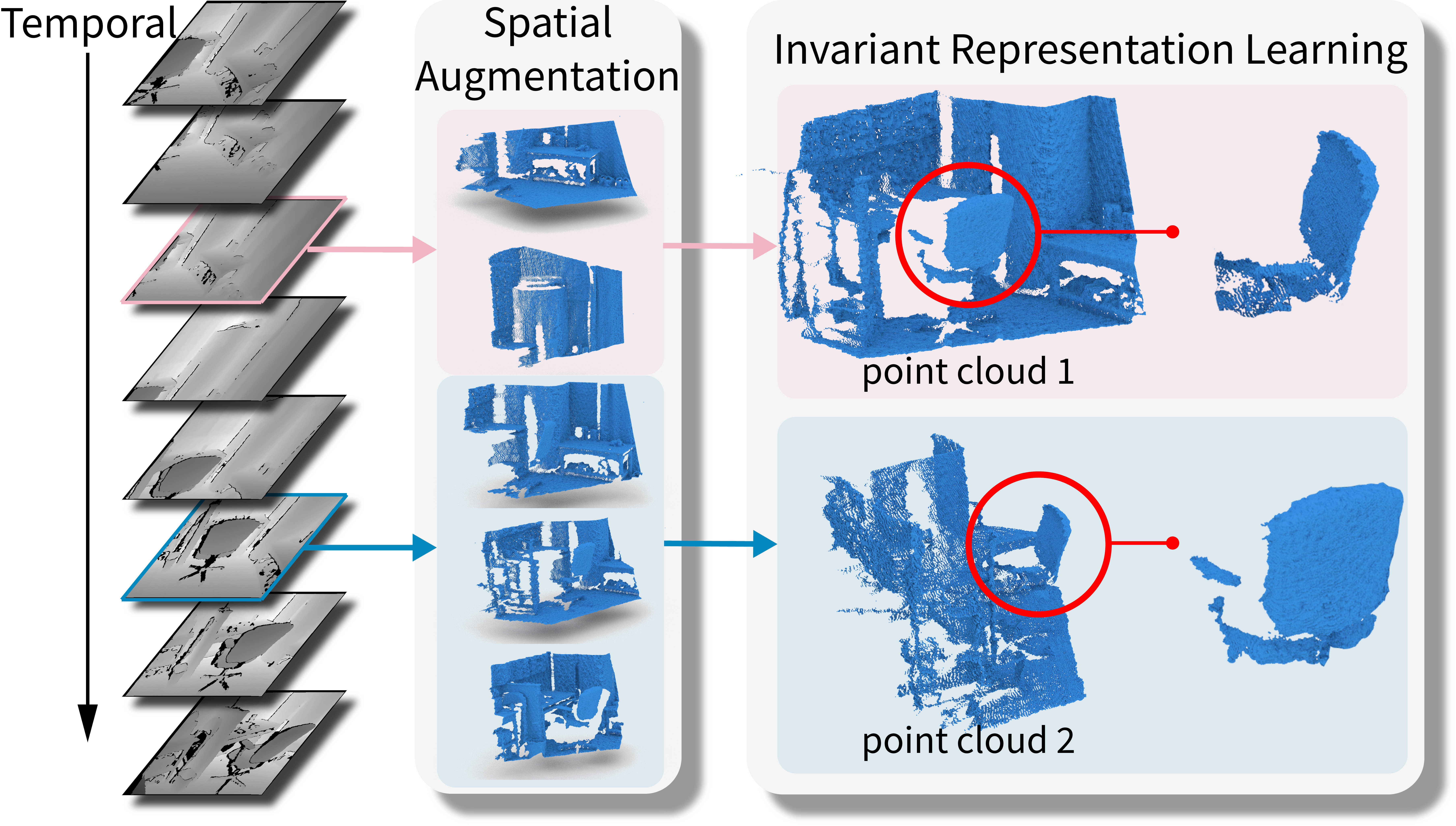}
    \caption{\textbf{Overview of our method.} By learning the spatio-temporal data invariance from a point cloud sequence, our method self-supevisedly learns an effective representation.}
    \label{fig:illustration}
\end{figure}

In this paper, we fill in the absence by exploiting self-supervised representation learning for 3D point clouds to address a long-standing problem in our community---the supervised training struggles at producing practical and generalizable pre-trained models due to the supervision-starved nature of the 3D data. Specifically, we consider the following three principles in model design and learning:

\paragraph{Simplicity}

Although self-supervised learning approaches for 3D point clouds exist, they rely exclusively on spatial analysis by reconstructing the 3D point clouds~\cite{achlioptas2017representation,yang2018foldingnet,sauder2019self,han2019view}. This static perspective of self-supervised learning is designed explicitly with complex operations, architectures, or losses, making it difficult to train and generalize to diversified downstream tasks. We believe such intricate designs are artificially introduced and unnecessary, and could be diminished or eliminated by complementing the missing \emph{temporal} contextual cues, akin to how infants may understand this world~\cite{gopnik2000scientist,smith2005development}.

\setstretch{0.97}

\paragraph{Invariance}

Learning data invariance via data augmentation and contrasting has shown promising results on images and videos~\cite{he2020momentum,chen2020simple,grill2020bootstrap}. A natural question arises: How could we introduce and leverage the invariance in 3D point clouds for self-supervised learning?
    
\paragraph{Generalizability}

Prior literature~\cite{achlioptas2017representation,yang2018foldingnet,sauder2019self,han2019view} has only verified the self-supervisedly learned representations in shape classification on synthetic datasets~\cite{chang2015shapenet}, which possesses dramatically different characteristics compared with the 3D data of natural indoor~\cite{song2015sun,dai2017scannet} or outdoor~\cite{geiger2012we} environments, thus failed to demonstrate sufficient generalizability to higher-level tasks (\eg, 3D object detection).

To adhere to the above principles and tackle the challenges introduced thereby, we devise a \acf{strl} framework to learn from \textit{unlabeled} 3D point clouds. Of note, \ac{strl} is remarkably \textbf{simple} by learning only from the positive pairs, inspired by the BYOL~\cite{grill2020bootstrap}. Specifically, \ac{strl} uses two neural networks, referred to as online and target networks, that interact and learn from each other. By augmenting one input, we train the online network to predict the target network representation of another \emph{temporally} correlated input, obtained by a separate augmentation process.

To learn the \textbf{invariant} representation~\cite{foldiak1991learning,wiskott2002slow}, we explore the inextricably spatio-temporal contextual cues embedded in 3D point clouds. In our approach, the online network's and target network's inputs are \emph{temporally} correlated, sampled from a point cloud sequence. Specifically, for natural images/videos, we sample two frames with a natural viewpoint change in depth sequences as the input pair. For synthetic data like 3D shape, we augment the original input by rotation, translation, and scaling to emulate the viewpoint change. The temporal difference between the inputs avails models of capturing the randomness and invariance across different viewpoints. Additional \emph{spatial} augmentations further facilitate the model to learn 3D spatial structures of point clouds; see examples in \cref{fig:illustration,sec:method}.

To \textbf{generalize} the learned representation, we adopt several practical networks as backbone models. By pre-training on large datasets, we verify that the learned representations can be readily adapted to downstream tasks directly or with additional feature fine-tuning. We also demonstrate that the learned representation can be generalized to distant domains, different from the pre-trained domains; \eg, the representation learned from ScanNet~\cite{dai2017scannet} can be generalized to shape classification tasks on ShapeNet~\cite{chang2015shapenet} and 3D object detection task on SUN RGB-D~\cite{song2015sun}.

We conduct \emph{extensive} experiments on various domains and test the performance by applying the pre-trained representation to downstream tasks, including 3D shape classification, 3D object detection, and 3D semantic segmentation. Next, we summarize our main findings.

\paragraph{Our method outperforms prior arts.}

By pre-training with \ac{strl} and applying the learned models to downstream tasks, it (i) outperforms the state-of-the-art unsupervised methods on ModelNet40~\cite{wu20153d} and reaches 90.9\% 3D shape classification accuracy with linear evaluation, (ii) shows significant improvements in semi-supervised learning with limited data, and (iii) boosts the downstream tasks by transferring the pre-trained models, \eg, it improves 3D object detection on SUN RGB-D~\cite{song2015sun} and KITTI dataset~\cite{geiger2012we}, and 3D semantic segmentation on S3DIS~\cite{armeni20163d} via fine-tuning.

\paragraph{Simple learning strategy leads to the satisfying performance of learned 3D representation.}

Through the ablative study in \cref{tab:ablation-spatial,tab:ablation-view}, we observe that \ac{strl} can learn the self-supervised representations with simple augmentations; it robustly achieves a satisfying accuracy (about 85\%) on ModelNet40 linear classification, which echoes recent findings~\cite{poursaeed2020self} that simply predicting the 3D orientation helps learn good representation for 3D point clouds.

\paragraph{The spatio-temporal cues boost the performance of learned representation.}

Relying on spatial or temporal augmentation alone only yield relatively low performance as shown in \cref{tab:ablation-spatial,tab:ablation-view}. In contrast, we achieve an improvement of 3\% accuracy by learning the invariant representations combining both spatial and temporal cues.

\paragraph{Pre-training on synthetic 3D shapes is indeed helpful for real-world applications.}

Recent study~\cite{xie2020pointcontrast} shows the representation learned from ShapeNet is not well-generalized to the downstream tasks. Instead, we report an opposite observation in \cref{tab:ablation-gap}, showing the representation pre-trained on ShapeNet can achieve comparable and even better performance while applying to downstream tasks that tackle complex data obtained in the physical world.

\section{Related Work}

\paragraph{Representation Learning on Point Clouds}

Unlike conventional representations of structured data (\eg, images), point clouds are unordered sets of vectors. This unique nature poses extra challenges to the learning of representations. Although deep learning methods on unordered sets~\cite{vinyals2015order,zaheer2017deep,maron2020learning} could be applied to point clouds~\cite{ravanbakhsh2016deep,zaheer2017deep}, these approaches do not leverage spatial structures.

Taking spatial structures into consideration, modern approaches like PointNet~\cite{qi2017pointnet} directly feed raw point clouds into neural networks; these networks ought to be permutation invariant as point clouds are unordered sets. PointNet achieves this goal by using the max-pooling operation to form a single feature vector representing the global context from a set of points. Since then, researches have proposed alternative representation learning methods with hierarchy~\cite{qi2017pointnet++,klokov2017escape,gadelha2018multiresolution}, convolution-based structure~\cite{hua2018pointwise,xu2018spidercnn,li2018pointcnn,su2018splatnet,zhang2019shellnet,wu2019pointconv,thomas2019kpconv}, or graph-based information aggregation~\cite{gadelha2018multiresolution,verma2018feastnet,shen2018mining,wang2019dynamic}. Operating directly on raw point clouds, these neural networks naturally provide per-point embedding, particularly effective for point-based tasks. Since the proposed \ac{strl} is flexible and compatible with various neural models serving as the backbone, our design of \ac{strl} leverages the efficacy introduced by per-point embedding.

\setstretch{1}

\paragraph{Unsupervised Representation Learning}

Unsupervised representation learning could be roughly categorized as either generative or discriminative approaches. Generative approaches typically attempt to reconstruct the input data in terms of pixel or point by modeling the distributions of data or the latent embedding. This process could be realized by energy-based modeling~\cite{lecun2006tutorial,ngiam2011learning,xie2016theory,gao2018learning,kumar2019maximum}, auto-encoding~\cite{vincent2008extracting,kingma2013auto,bengio2013generalized}, or adversarial learning~\cite{goodfellow2014generative}. However, this unsupervised mechanism is computationally expensive, and the learning of generalizable representation unnecessarily relies on recovering such high-level details.

Discriminative approaches, including self-supervised learning, unsupervisedly generate discriminative labels to facilitate representation learning, recently achieved by various contrastive mechanisms~\cite{he2020momentum,oord2018representation,hjelm2018learning,henaff2019data,bachman2019learning,tian2019contrastive,tian2020makes}. Different from generative approaches that maximize the data likelihood, recent contrastive approaches maximally preserve the mutual information between the input data and its encoded representation. Following BYOL~\cite{grill2020bootstrap}, we exclude negative pairs in contrastive learning and devise \ac{strl} to construct a stable and invariant representation through a moving average target network.

\paragraph{Self-supervised Learning of Point Clouds}

Although various approaches~\cite{wu2016learning,achlioptas2017representation,gadelha2018multiresolution,yang2018foldingnet,li2018point,zhao20193d,sun2020pointgrow,poursaeed2020self} have been proposed for unsupervised learning and generation of point clouds, these approaches have merely demonstrated efficacy in shape classification tasks on synthetic datasets while ignoring higher-level tasks of pre-trained models on natural 3D scenes. More recent work starts to demonstrate the potentials for high-level tasks such as 3D object detection and 3D semantic segmentation. For instance, Sauder \etal~\cite{sauder2019self} train the neural network to reconstruct point clouds with self-supervised labels generated by randomly arranging object parts, and Xie \etal~\cite{xie2020pointcontrast} learn from dense correspondences between different views with a contrastive loss. In comparison, the proposed \ac{strl} is much simpler without computing the dense correspondences or reconstruction loss; it relies solely on spatio-temporal contexts and structures of point clouds, yielding more robust and improved performances on various high-level downstream tasks.

\begin{figure}[t!]
    \centering
    \includegraphics[width=\linewidth]{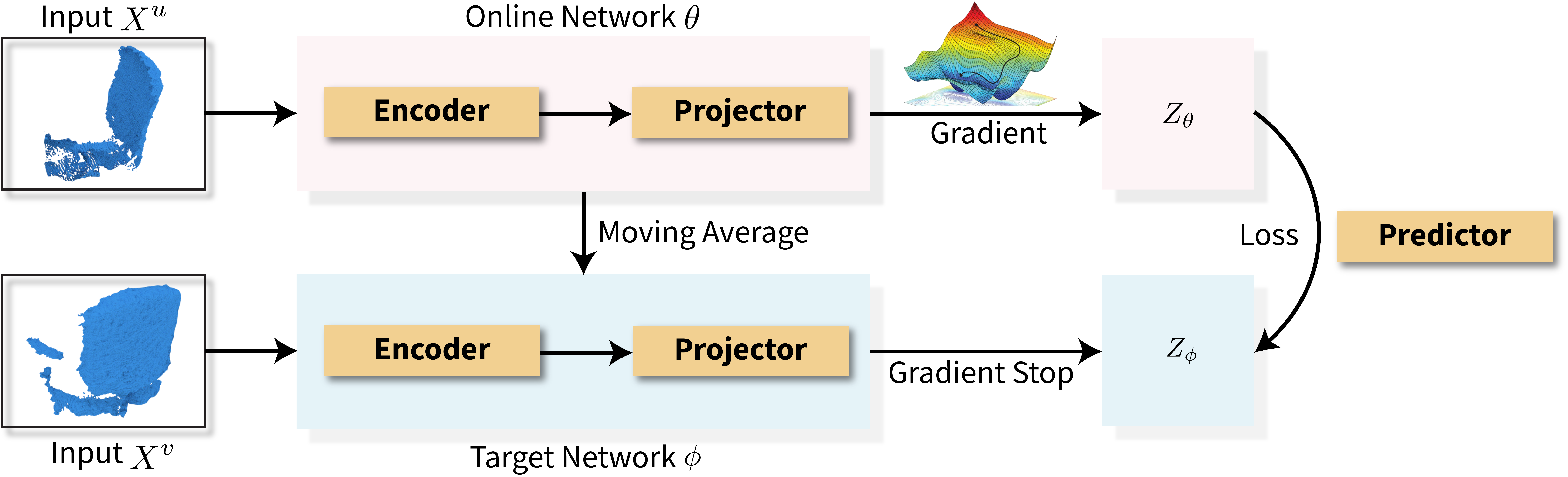}
    \caption{\textbf{Illustration of our self-supervised learning framework.} Given two spatio-temporal correlated 3D point clouds, the online network predicts the target network's representation via a predictor. Parameters of the target network are updated by the online network's moving average.}
    \label{fig:framework}
\end{figure}

\section{Spatio-temporal Representation Learning}\label{sec:method}

We devise the proposed \acf{strl} based on BYOL~\cite{grill2020bootstrap} and extend its simplicity to the learning of 3D point cloud representation. \cref{fig:framework} illustrates the proposed method.

\subsection{Building Temporal Sequence of Point Clouds}\label{sec:sequence}

To learn a simple, invariant, and generalizable representation for 3D point clouds, we formulate the representation learning as training with sequences of potentially partial and cluttered 3D point clouds of objects or scenes. Given a sequence of potentially non-uniformly sampled time steps, we denote the corresponding point cloud sequence as $\mathcal{P} = \left\{p_t\right\}_{t=1}^T$. We devise two approaches to generating the training point cloud sequences to handle various data sources.

\paragraph{Natural Sequence}

Natural sequences refer to the data sequences captured by RGB-D sensors, wherein each depth image $I_t$ is a projected view of the scene. Given the camera pose (extrinsic parameters) at each time step $c^{ex}_t$, we back-project depth images with intrinsic parameters $c^{in}$ and obtain a sequence of point clouds $\{p_t\}$ in world coordinate:
\begin{equation}\label{eq:project}
    p_t = Backproj(I_t, c^{ex}_t, c^{in}), \quad{} t=1, \cdots, T.
\end{equation}

\paragraph{Synthetic Sequence}

Static point clouds are intrinsically spatial, missing the crucial temporal dimension compared to natural sequences. Given a point cloud $p_0$, we solve this problem by generating a synthetic sequence. Specifically, we consecutively rotate, translate, and scale the original point cloud to construct a sequence of point clouds $\{p_t\}$:
\begin{equation}
    p_t = R_t (p_{t-1}), \quad{} t=1, \cdots, T,
    \label{eq:syn}
\end{equation}
where $t$ is the index of transformations, and $R_t$ the sampled transformation, emulating temporal view changes.

\subsection{Representation Learning}

We design \ac{strl} to unsupervisedly learn the representations through the interactions of two networks: the \textbf{online} network and \textbf{target} network. Here, the essence of self-supervised learning is to train the online network to accurately predict the target network's representation.

Specifically, the online network parameterized by $\theta$ consists of two components: a backbone encoder $e_\theta$ and a feature projector $f_\theta$. Similarly, the target network parameterized by $\phi$ has a backbone encoder $e_\phi$ and feature projector $f_\phi$. In addition, a predictor $r$ with parameters regresses the target presentation: The target network serves as regression targets to train the online network, and its parameters $\phi$ are an exponential moving average of the online parameters $\theta$,
\begin{equation}
    \phi \leftarrow \tau \phi + (1 - \tau) \theta,
    \label{eq:average}
\end{equation}
where $\tau \in [0,1]$ is the decay rate of the moving average.

\setstretch{0.99}

Given a sequence of point clouds $\mathcal{P}$, we sample two frames of point clouds $p^u, p^v \in \mathcal{P}$ by a temporal sampler $\mathcal{T}$. With a set of spatial augmentations $\mathcal{A}$ (see details in \cref{sec:detail}), \ac{strl} generates two inputs $x^u = a^u(p^u)$ and $x^v = a^v(p^v)$, where $a^u, a^v \in \mathcal{A}$. For each input, the online network and target network generate $z_\theta = f_\theta(e_\theta(x^u)), z_\phi = f_\phi(e_\phi(x^v))$, respectively. With the additional predictor $r$, the goal of \ac{strl} is to minimize the mean squared error between the normalized predictions and target projections:
\begin{equation}
    \small
    \mathcal{L}_{u\rightarrow v} = \left \| \frac{r(z_\theta)}{\left\|r(z_\theta)\right\|_2} -  \frac{z_\phi}{\left\|z_\phi\right\|_2}\right\|^2_2 = 2 - 2 \cdot \frac{\left \langle r(z_\theta), z_\phi \right\rangle}{\left\|r(z_\theta)\right\|_2 \cdot \left\|z_\phi\right\|_2}
    \label{eq:loss}
\end{equation}

Finally, we symmetrize the loss in \cref{eq:loss} to compute $\mathcal{L}_{v \rightarrow u}$ by separately feeding $x^v$ to the online network and $x^u$ to the target network. The total loss is defined as:
\begin{equation}
    \mathcal{L}_{total} = \mathcal{L}_{u \rightarrow v} + \mathcal{L}_{v \rightarrow u}. 
\end{equation}

Within each training step, only the parameters of the online network and predictor are updated. The target network's parameters are updated after each training step by \cref{eq:average}. Similar to~\cite{he2020momentum,grill2020bootstrap}, we only keep the backbone encoder of the online network $e_\theta$ at the end of the training as the learned model. \cref{alg:strl} details the proposed \ac{strl}.

\begin{algorithm}[ht!]
    \caption{\ac{strl} of 3D point clouds}
    \label{alg:strl}
    \small
    \SetAlgoLined
    \KwInput{}
    $\left\{\mathcal{P}\right\}$: a set of 3D point cloud sequences\;
    $\mathcal{T}, \mathcal{A}$: temporal sampler and spatial augmentations\;
    $e_\theta, f_\theta$: online encoder and projector with parameter $\theta$\;
    $e_\phi, f_\phi$: target encoder and projector with parameter $\phi$\;
    $r$: predictor\;
    $K$: number of optimization steps\;
    $N$: batch size.\\
    \KwOutput{online encoder $e_\theta$.}
    \For{$k=1$ \KwTo $K$}{
        \tcc{sample batches of temporal-correlated point clouds}
        $\mathcal{B} \leftarrow \left\{p^u_i, p^v_i \in \mathcal{T}(\left\{\mathcal{P}\right\}) \right\}_{i=1}^{N}$\\
        \For{$i=1$ \KwTo $N$}{
            \tcc{sample spatial augmentations}
            $a^u, a^v \in \mathcal{A}$\\
            \tcc{generate inputs}
            $x^u = a^u(p^u), x^t = a^v(p^v)$\\
            \tcc{project}
            $z_\theta = f_\theta(g_\theta(x^u)), z_\phi = f_\phi(g_\phi(x^v))$\\
            \tcc{compute loss}
            $\mathcal{L}_{u\rightarrow v} = - 2 \cdot \frac{\left \langle r(z_\theta), z_\phi \right\rangle}{\left\|r(z_\theta)\right\|_2 \cdot \left\|z_\phi\right\|_2}$\\
            \tcc{compute total \& symmetric loss}
            $\mathcal{L}_{total} = \mathcal{L}_{u \rightarrow v} + \mathcal{L}_{v \rightarrow u}$
        }
        \tcc{update online network \& predictor}
        $\theta, r = optimize(\theta, r, \mathcal{L}_{total})$\\
        \tcc{update target network}
        $\phi \leftarrow \tau \phi + (1 - \tau) \theta$
    }
\end{algorithm}

\setstretch{0.96}

\section{Implementation Details}\label{sec:detail}

\paragraph{Synthetic Sequence Generation}

\begin{figure*}[t!]
    \centering
    \includegraphics[width=\linewidth]{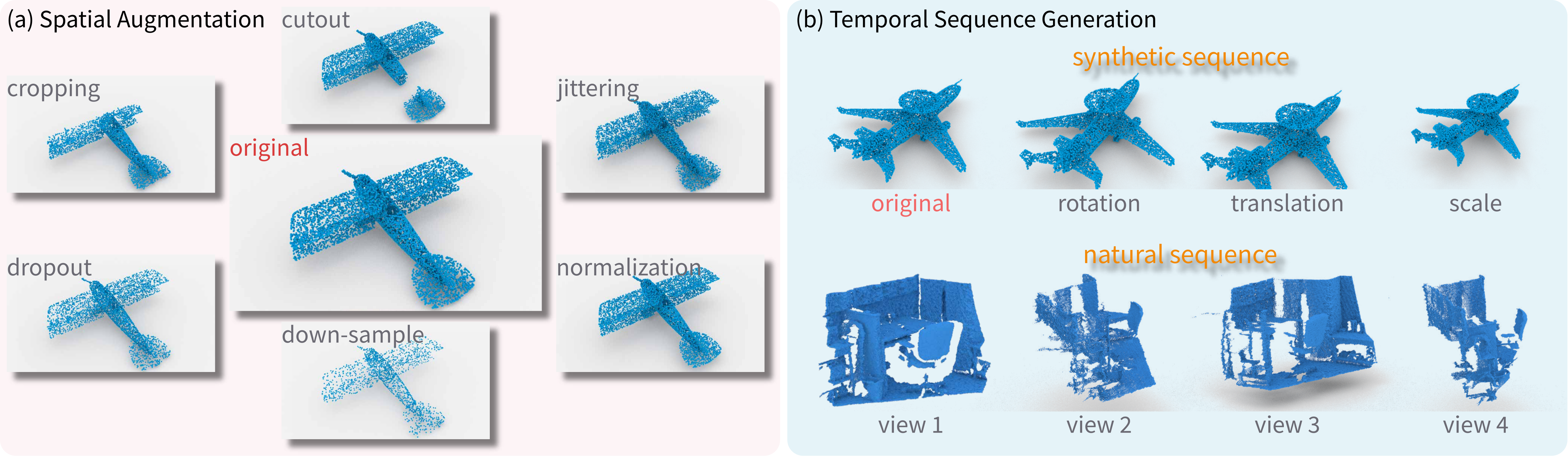}
    \caption{\textbf{Spatial data augmentation and temporal sequence generation.} Except for the natural sequence generation, each type of augmentation transforms the input point cloud data stochastically with certain internal parameters.}
    \label{fig:augmentation}
\end{figure*}

We sample the combination of following transformations to construct the function $R()$ in \cref{eq:syn}; see an illustration in \cref{fig:augmentation}b:
\begin{itemize}[leftmargin=*,noitemsep,nolistsep]
    \item Random rotation. For each axis, we draw random angles within $15^\circ$ and rotate around it. 
    \item Random translation. We translate the point cloud globally within 10\% of the point cloud dimension.
    \item Random scaling. We scale the point cloud with a factor $s\in[0.8, 1.25]$.
\end{itemize}

To further increase the randomness, each transformation is sampled and applied with a probability of $0.5$.

\paragraph{Spatial Augmentation}

The spatial augmentation transforms the input by changing the point cloud's local geometry, which helps \ac{strl} to learn a better spatial structure representation of point clouds. Specifically, we apply the following transformations, similar to the image data augmentation; see an illustration in \cref{fig:augmentation}a.
\begin{itemize}[leftmargin=*,noitemsep,nolistsep]
    \item Random cropping. A random 3D cuboid patch is cropped with a volume uniformly sampled between 60\% and 100\% of the original point cloud. The aspect ratio is controlled within $[0.75, 1.33]$.
    \item Random cutout. A random 3D cuboid is cut out. Each dimension of the 3D cuboid is within $[0.1, 0.4]$ of the original dimension. 
    \item Random jittering. Each point's 3D locations are shifted by a uniformly random offset within $[0, 0.05]$. 
    \item Random drop-out. We randomly drop out 3D points by a drop-out ratio within $[0, 0.7]$.
    \item Down-sampling. We down-sample point clouds based on the encoder's input dimension by randomly picking the necessary amount of 3D points.
    \item Normalization. We normalize the point cloud to fit a unit sphere while training on synthetic data~\cite{chang2015shapenet}.
\end{itemize}

Among these augmentations, cropping and cutout introduce more evident changes to the point clouds' spatial structures. As such, we apply them with a probability of $0.5$. 

\paragraph{Training}

We use the LARS optimizer~\cite{you2017scaling} with a cosine decay learning rate schedule~\cite{loshchilov2016sgdr}, with a warm-up period of 10 epochs but without restarts. For the target network, the exponential moving average parameter starts with $\tau_{start}= 0.996$ and is gradually increased to $1$ during the training. Specifically, we set $\tau = 1 - (1 - \tau_{start}) \cdot (cos(\pi k / K) + 1) / 2$ with $k$ being the current training step and $K$ the maximum number of training steps.

\ac{strl} is favorable and generalizable to different backbone encoders; see details about the encoder structure for each specific experiment in \cref{sec:exp}. The projector and predictor are implemented as multi-layer perceptions (MLPs) with activation~\cite{nair2010rectified} and batch normalization~\cite{ioffe2015batch}; see the \emph{supplementary materials} for detailed network structures. We use a batch size ranging from 64 to 256 split over 8 TITAN RTX GPUs for most of the pre-trained models.

\section{Experiment}\label{sec:exp}

We start by introducing how to pre-train \ac{strl} on various data sources in \cref{sec:dataset}. Next, we evaluate these pre-trained models on various downstream tasks in \cref{sec:downstream}. At length, in \cref{sec:ablation}, we analyze the effects of different modules and parameters in our model, with additional analytic experiments and discussions of open problems.

\subsection{Pre-training}\label{sec:dataset}

To recap, as detailed in \cref{sec:sequence}, we build the sequences of point clouds and perform the pre-training of \ac{strl} to learn the spatio-temporal invariance of point cloud data. For synthetic shapes and natural indoor/outdoor scenes, we generate temporal sequences of point clouds and sample input pairs using different strategies detailed below.

\subsubsection{Synthetic Shapes}

\paragraph{ShapeNet}

We learn the self-supervised representation model from the ShapeNet~\cite{chang2015shapenet} dataset. It consists of 57,448 synthetic objects from 55 categories. We pre-process the point clouds following Yang \etal~\cite{yang2018foldingnet}. By augmenting each point cloud into two different views with temporal transformations defined in \cref{eq:syn}, we generate two temporal-corrected point clouds. The spatial augmentations are further applied to produce the pair of point clouds as the input.

\subsubsection{Natural Indoor and Outdoor Scenes}

We also learn the self-supervised representation models from natural indoor and outdoor scenes, in which sequences of point clouds are readily available. Using RGB-D sensors, sequences of depth images can be captured by scanning over different camera poses. Since most scenes are captured smoothly, we learn the temporal invariance from the temporal correlations between the adjacent frames.

\paragraph{ScanNet}

For indoor scenes, we pre-train on the ScanNet dataset~\cite{dai2017scannet}. It consists of 1,513 reconstructed meshes for 707 unique scenes. In experiments, we find that increasing the frame-sampling frequency only makes a limited contribution to the performance. Hence, we sub-sample the raw depth sequences every 100 frames as the keyframes for each scene, resulting in 1,513 sequences and roughly 25 thousand frames in total. During pre-training, we generate fixed-length sliding windows based on the keyframes of each sequence and sample two random frames within each window. By back projecting the two frames with \cref{eq:project}, we generate point clouds in the world coordinate. We use the camera position to translate the two point clouds into the same world coordinate; the camera center of the first frame is the origin. 

\paragraph{KITTI}

For outdoor scenes, we pre-train on the KITTI dataset~\cite{geiger2013vision}. It includes 100+ sequences divided into 6 categories. For each scene, images and point clouds are recorded at roughly 10 FPS. We only use the sequences of point clouds captured by the Velodyne lidar sensor. On average, each frame has about 120,000 points. Similar to ScanNet, we sub-sample the keyframes and sample frame pairs within sliding windows as training pairs. 

For pre-training on natural scenes, we further enhance the data diversity by applying synthetic temporal transformations in \cref{eq:syn} to the two point clouds. At length, the spatial data augmentation is applied to both point clouds.

\subsection{Downstream Tasks}\label{sec:downstream}

For each downstream task below, we present the model structures, experimental settings, and results. Please refer to the \emph{supplementary materials} for training details.

\subsubsection{Shape Understanding}\label{sec:shape_cls}

We adopt the protocols presented in prior work~\cite{achlioptas2017representation,sauder2019self,wu2016learning,yang2018foldingnet} to evaluate the shape understanding capability of our pre-trained model using the ModelNet40~\cite{wu20153d} benchmark. It contains 12,331 objects (9,843 for training and 2,468 for testing) from 40 categories. We pre-process the data following Qi \etal~\cite{qi2017pointnet}, such that each shape is sampled to 10,000 points in unit space.

As detailed in \cref{sec:dataset}, we pre-train the backbone models on ShapeNet dataset. We measure the learned representations using the following evaluation metrics.

\setstretch{0.98}

\paragraph{Linear Evaluation for Shape Classification}

To classify 3D shapes, We append a linear Support Vector Machine (SVM) on top of the encoded global feature vectors. Following Sauder \etal~\cite{sauder2019self}, these global features are constructed by extracting the activation after the last pooling layer. Our \ac{strl} is flexible to work with various backbones; we select two practical ones---PointNet~\cite{qi2017pointnet} and DGCNN~\cite{wang2019dynamic}. The SVM is trained with the extracted global features from the training sets of ModelNet40 datasets. We randomly sample 2,048 points from each shape during both pre-training and SVM training. 

\cref{tab:svmres} tabulates the classification results on test sets. The proposed \ac{strl} outperforms all the state-of-the-art unsupervised and self-supervised methods on ModelNet40.

\begin{table}[htb!]
    \centering
    \caption{\textbf{Comparisons of the linear evaluation for shape classification on ModelNet40}. A linear classifier is trained on the representations learned by different self-supervised approaches on the ShapeNet dataset.}
    \label{tab:svmres}
    \scalebox{0.85}{%
        \begin{tabular}{lc}
            \toprule
            Method & ModelNet40\\
            \midrule
            3D-GAN~\cite{wu2016learning} & 83.3\% \\
            Latent-GAN~\cite{achlioptas2017representation} & 85.7\% \\
            SO-Net~\cite{2018SO} & 87.3\%\\
            FoldingNet~\cite{yang2018foldingnet} & 88.4\%\\
            MRTNet~\cite{han2019multi} & 86.4\% \\
            3D-PointCapsNet~\cite{yang2018foldingnet} & 88.9\%\\
            MAP-VAE~\cite{yang2018foldingnet} & 88.4\%\\
            Sauder \etal + PointNet~\cite{sauder2019self} & 87.3\% \\
            Sauder \etal + DGCNN~\cite{sauder2019self} & 90.6\% \\
            Poursaeed \etal + PointNet~\cite{poursaeed2020self} & 88.6\% \\
            Poursaeed \etal + DGCNN~\cite{poursaeed2020self} & 90.7\%  \\
            \midrule 
            \ac{strl} + PointNet (ours) & 88.3\% \\
            \ac{strl} + DGCNN (ours) & \textbf{90.9\%}\\
            \bottomrule 
        \end{tabular}%
    }%
\end{table}

\paragraph{Supervised Fine-tuning for Shape Classification}

We also evaluate the learned representation models by supervised fine-tuning. The pre-trained model serves as the point cloud encoder's initial weight, and we fine-tune the DGCNN network given the labels on the ModelNet40 dataset. Our \ac{strl} leads to a marked performance improvement of up to 0.9\% on the final classification accuracy; see \cref{tab:dgcnncls_full}. This improvement is more significant than previous methods; it even attains a comparable performance set by the state-of-the-art supervised learning method~\cite{zhang2019shellnet}.

\begin{table}[htb!]
    \centering
    \caption{\textbf{Shape classification fine-tuned on ModelNet40}. The self-supervised pre-trained model serves as the initial weight for supervised learning methods.}
    \label{tab:dgcnncls}
    \begin{subtable}[h]{\linewidth}
        \centering
        \caption{Fine-tuned on Full Training Set}
        \label{tab:dgcnncls_full}
        \scalebox{0.85}{%
            \begin{tabular}{llc}
                \toprule
                Category          & Method & Accuracy \\
                \midrule
                \multirow{5}{*}{\emph{Supervised}}
                & PointNet~\cite{qi2017pointnet} & 89.2\%\\
                & PointNet++~\cite{qi2017pointnet++} & 90.7\%\\
                & PointCNN~\cite{li2018pointcnn} & 92.2\%\\
                & DGCNN~\cite{wang2019dynamic} & 92.2\%\\
                & ShellNet~\cite{zhang2019shellnet} & 93.1\%\\
                \midrule
                \multirow{2}{*}{\emph{Self-supervised}}
                & Sauder \etal + DGCNN~\cite{sauder2019self} & 92.4\% \\
                & \ac{strl} + DGCNN (ours) & \textbf{93.1\%}\\
                \bottomrule
            \end{tabular}%
        }%
    \end{subtable}%
    \\%
    \begin{subtable}[h]{\linewidth}
        \centering
        \caption{Fine-tuned on Few Training Samples}
        \label{tab:dgcnncls_few}
        \scalebox{0.85}{%
            \begin{tabular}{lcccc}
                \toprule
                Method &  1\% & 5\%  & 10\% & 20\% \\
                \midrule
                DGCNN & 58.4\% & 80.7\% & 85.2\% & 88.1\%\\
                \ac{strl} + DGCNN & \textbf{60.5}\% & \textbf{82.7}\% & \textbf{86.5}\% & \textbf{89.7}\%\\
                \bottomrule 
            \end{tabular}%
        }%
    \end{subtable}
\end{table}

Furthermore, we show that our pre-trained models can significantly boost the classification performance in semi-supervised learning where limited labeled training data is provided. Specifically, we randomly sample the training data with different proportions and ensure at least one sample for each category is selected. Next, we fine-tune the pre-trained model on these limited samples with supervision and evaluate its performance on full test sets. \cref{tab:dgcnncls_few} summarizes the results measured by accuracy. It shows that the proposed model obtains 2.1\% and 1.6\% performance gain when 1\% and 20\% of the training samples are available; our self-supervised models would better facilitate downstream tasks when fewer training samples are available.

\setstretch{0.99}

\paragraph{Embedding Visualization}

\begin{figure}[t!]
    \centering
    \includegraphics[width=\linewidth]{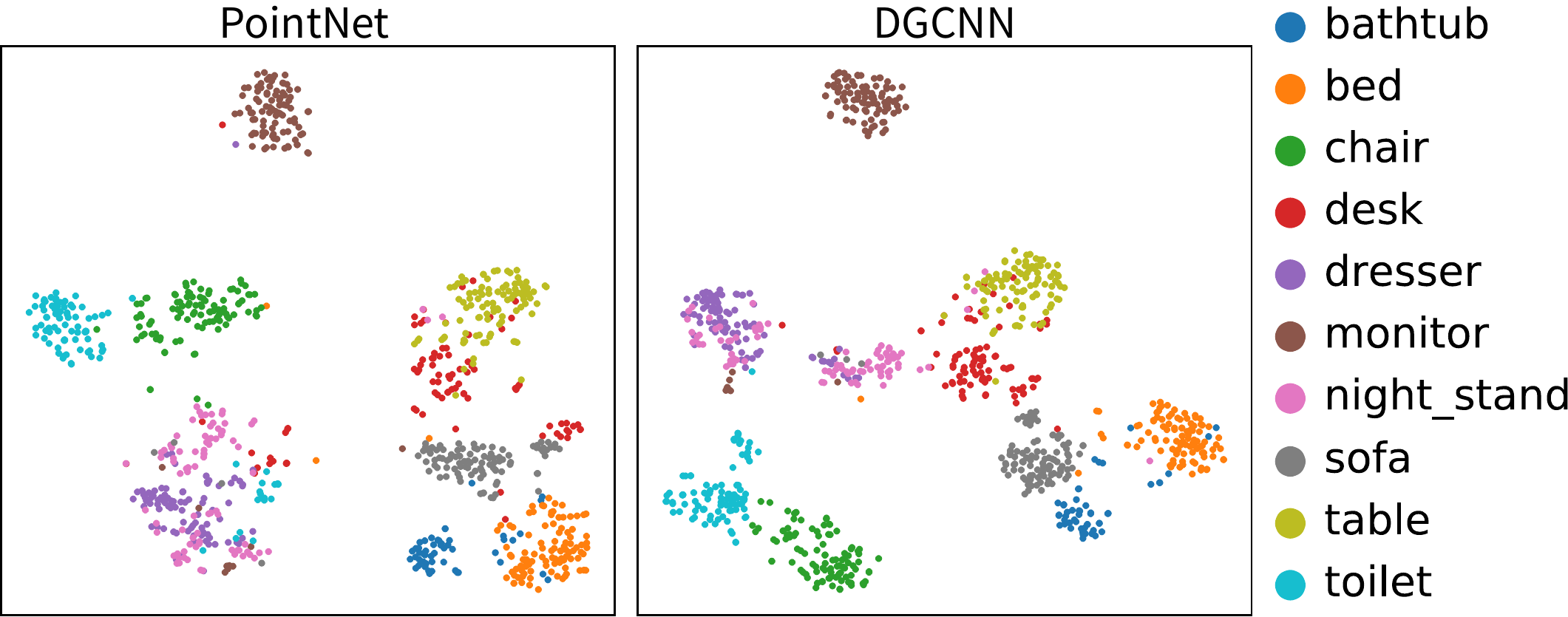}
    \caption{\textbf{Visualization of learned features}. We visualize the extracted features for each sample in ModelNet10 test set using t-SNE. Both models are pre-trained on ShapeNet.}
    \label{fig:emb_vis}
\end{figure}

We visualize the learned features of PointNet and DGCNN model with our self-supervised method in \cref{fig:emb_vis}; it displays the embedding for samples of different categories in the ModelNet10 test set. t-SNE~\cite{maaten2008visualizing} is adopted for dimension reduction. We observe that both pre-trained models well separate most samples based on categories, except \emph{dressers} and \emph{night stands}; they usually look similar and are difficult to distinguish.

\subsubsection{Indoor Scene Understanding}\label{sec:indoor}

Our proposed \ac{strl} learns representations based on view transformation, suitable for both synthetic shapes and natural scenes. Consequently, unlike prior work that primarily performs transfer learning to shape understanding, our method can also boost the indoor/outdoor scene understanding tasks. We start with the indoor scene understanding in this section. We first pre-train our \ac{strl} self-supervisedly on the ScanNet dataset as described in \cref{sec:dataset}. Next, we evaluate the performance of 3D object detection and semantic segmentation through fine-tuning with labels.

\paragraph{3D Object Detection}

3D object detection requires the model to predict the 3D bounding boxes with their object categories based on the input 3D point cloud. After pre-training, we fine-tune and evaluate the model on the SUN RGB-D~\cite{song2015sun} dataset. It contains 10,335 single-view RGB-D images, split into 5,285 training samples and 5,050 validation samples. Objects are annotated with 3D bounding boxes and category labels. We conduct this experiment with VoteNet~\cite{qi2019deep}, which is a widely-used model with 3D point clouds as input. During pre-training, we slightly modify its PointNet++~\cite{qi2017pointnet++} backbone by adding a max-pooling layer at the end to obtain the global features. \cref{tab:obj-det} summarizes the results. The pre-training improves the detection performance by 1.2 mAP compared against training VoteNet from scratch, demonstrating that the representation learned from a large dataset, \ie, ScanNet, can be successfully transferred to a different dataset and improve the performances of high-level tasks via fine-tuning. It also outperforms the state-of-the-art self-supervised learning method~\cite{xie2020pointcontrast} by 0.7 mAP.\footnote{The model pre-trained on ShapeNet achieves better results as 59.2 mAP, which is analyzed and explained in \cref{sec:ablation}}

\begin{table}[htb!]
    \centering
    \caption{\textbf{3D object detection fine-tuned on SUN RGB-D}}
    \label{tab:obj-det}
    \scalebox{0.85}{%
        \begin{tabular}{lccc}
            \toprule
            Model & Method & Input & mAP@0.25 IoU\\
            \midrule
            \multirow{2}{*}{VoteNet} & \multirow{2}{*}{\emph{from scratch}}  & Geo+Height  & 57.7\\
            &  & Geo & 57.0\\
            \midrule
            {SR-UNet~\cite{choy20194d}} & PointContrast~\cite{xie2020pointcontrast} & Geo & 57.5\\
            {VoteNet} & \ac{strl} (ours) & Geo & \textbf{58.2}\\
            \bottomrule
        \end{tabular}%
    }%
\end{table}

\paragraph{3D Semantic Segmentation}

We transfer the pre-trained model to the 3D semantic segmentation task on the Stanford Large-Scale 3D Indoor Spaces (S3DIS)~\cite{armeni20163d} dataset. This dataset contains 3D point clouds scanned from 272 rooms in 6 indoor areas, with each point annotated into 13 categories. We follow the setting in Qi \etal~\cite{qi2017pointnet} and Wang \etal~\cite{wang2019dynamic} and split each room into $1m\times1m$ blocks. Different from them, we use 4,096 points with only geometric features (XYZ coordinates) as the model input. In this experiment, the DGCNN network is firstly pre-trained on ScanNet with \ac{strl}. Here, we focus on semi-supervised learning with only limited labeled data. As such, we fine-tune the pre-trained model on one area in Area 1-5 each time and test the model on Area 6. As shown in \cref{tab:sem-seg}, the pre-trained models consistently outperform the models trained from scratch, especially with a small training set.

\begin{table}[htb!]
    \centering
    \caption{\textbf{3D semantic segmentation fine-tuned on S3DIS}. We train the pre-trained or initialized models in a semi-supervised manner on one of the Areas 1-5. Performances below are evaluated on Area 6 of the S3DIS dataset.}
    \label{tab:sem-seg}
    \scalebox{0.85}{%
        \begin{tabular}{cccc}
            \toprule
            Fine-tuning Area & Method & Acc. & mIoU\\
            \midrule
            \multirow{2}{*}{Area 1 (3687 \emph{samples})} & \emph{from scratch} & 84.57\% & 57.85\\
            & \ac{strl} & \textbf{85.28\%} & \textbf{59.15} \\
            \midrule
            \multirow{2}{*}{Area 2 (4440 \emph{samples})} & \emph{from scratch} & 70.56\% & 38.86 \\
            & \ac{strl} & \textbf{72.37\%} & \textbf{39.21} \\
            \midrule
            \multirow{2}{*}{Area 3 (1650 \emph{samples})} & \emph{from scratch} & 77.68\% & 49.49\\
            & \ac{strl} & \textbf{79.12\%} & \textbf{51.88} \\
            \midrule
            \multirow{2}{*}{Area 4 (3662 \emph{samples})} & \emph{from scratch} & 73.55\% & 38.50\\
            & \ac{strl} & \textbf{73.81\%} & \textbf{39.28} \\
            \midrule
            \multirow{2}{*}{Area 5 (6852 \emph{samples})} & \emph{from scratch} & 76.85\% & 48.63 \\
            & \ac{strl} & \textbf{77.28\%} & \textbf{49.53} \\
            \bottomrule 
        \end{tabular}%
    }%
\end{table}

\subsubsection{Outdoor Scene Understanding}

Compared with indoor scenes, point clouds captured in outdoor environments are much sparser due to the long-range nature of Lidar sensors, posing additional challenges. In this section, we evaluate the performance of the proposed \ac{strl} by transferring the learned visual representations to the 3D object detection task for outdoor scenes.

As described in \cref{sec:dataset}, we pre-train the model on the KITTI dataset with PV-RCNN~\cite{shi2020pv}---the state-of-the-art model for 3D object detection. Similar to VoteNet, we modify the backbone network of PV-RCNN for pre-training by adding a max-pooling layer to obtain the global features.

\setstretch{1}

We fine-tune the pre-trained model on KITTI 3D object detection benchmark~\cite{geiger2012we}, a subset of the KITTI raw data. In this benchmark, each point cloud is annotated with 3D object bounding boxes. The subset includes 3,712 training samples, 3,769 validation samples, and 7,518 test samples. \cref{tab:obj-det-outdoor} tabulates results. On all three categories, models pre-trained with \ac{strl} outperform the model trained from scratch. In particular, for the cyclist category where the least training samples are available, the proposed \ac{strl} generates a marked performance elevation. We further freeze the backbone model while fine-tuning; the results reveal that models with the pre-trained backbone reach a comparable performance compared with training models from scratch.

\begin{table}[htb!]
    \centering
    \caption{\textbf{3D object detection fine-tuned on KITTI}. We report 3D detection performance with moderate difficulty on the val set of KITTI dataset. Performances below are evaluated by mAP with 40 recall positions.}
    \label{tab:obj-det-outdoor}
    \scalebox{0.85}{%
        \begin{tabular}{m{27mm}cccccc}
            \toprule
            \multirow{2}{*}{Method} & \multicolumn{2}{c}{Car (IoU=0.7)} & \multicolumn{2}{c}{Pedestrian} & \multicolumn{2}{c}{Cyclist} \\
            & 3D & BEV & 3D & BEV & 3D & BEV \\
            \midrule
            PV-RCNN \quad{}\quad{}(\emph{from scratch}) & 84.50 & 90.53 & 57.06 & 59.84 & 70.14 & 75.04 \\
            \ac{strl} + PV-RCNN (\emph{frozen backbone}) & 81.63 & 87.84 & 39.62 & 42.41 & 69.65 & 74.20 \\
            \ac{strl} + PV-RCNN & \textbf{84.70} & \textbf{90.75} & \textbf{57.80} & \textbf{60.83} &\textbf{71.88} & \textbf{76.65} \\
            \bottomrule
        \end{tabular}%
    }%
\end{table}
 
\subsection{Analytic Experiments and Discussions}\label{sec:ablation}

\paragraph{Generalizability: ScanNet vs ShapeNet Pre-training}

What kind of data would endow the learned model with better generalizability to other data domains remains an open problem in 3D computer vision. To elucidate this problem, we pre-train the model on the existing largest natural dataset ScanNet and synthetic data ShapeNet, and test their generalizability to different domains. \cref{tab:ablation-gap} tabulates our cross-domain experimental settings and results, demonstrating the successful transfer from models pre-trained on natural scenes to synthetic shape domain, achieving comparable shape classification performance under linear evaluation.

\begin{table}[htb!]
    \centering
    \caption{\textbf{Ablation study: cross-domain generalizability}}
    \label{tab:ablation-gap}
    \begin{subtable}[h]{\linewidth}
        \centering
        \caption{Linear evaluation for shape classification on ModelNet40.}
        \scalebox{0.85}{%
            \begin{tabular}{lcc}
                \toprule
                Method & Pre-train Dataset & Accuracy \\
                \midrule
                \multirow{2}{*}{\ac{strl} + DGCNN (linear)} & ScanNet & 90.4\% \\
                & ShapeNet & \textbf{90.9\%}\\
                \midrule
                \multirow{2}{*}{\ac{strl} + DGCNN (fine-tune)} & ScanNet & 92.9\%\\
                & ShapeNet & \textbf{93.1\%}\\
                \bottomrule 
            \end{tabular}%
        }%
    \end{subtable}%
    \\%
    \begin{subtable}[h]{\linewidth}
        \centering
        \caption{Fine-tuned 3D object detection on SUN RGB-D.}
        \scalebox{0.85}{%
            \begin{tabular}{lcc}
                \toprule
                Method & Pre-train Dataset & mAP@0.25 IoU\\
                \midrule
                \multirow{2}{*}{\ac{strl} + VoteNet} & ScanNet & 58.2\\
                & ShapeNet & \textbf{59.2}\\
                \bottomrule 
            \end{tabular}%
        }%
    \end{subtable}
\end{table}

\setstretch{0.99}

Additionally, we report an opposite observation in contrast to a recent study~\cite{xie2020pointcontrast}. Specifically, the VoteNet model pre-trained on the ShapeNet dataset achieves better performance than ScanNet pre-training in SUN RGB-D object detection, demonstrating better generalizability of ShapeNet data. We believe three potential reasons lead to such conflicting results: (i) The encoder adapted to learn the point cloud features in Xie \etal~\cite{xie2020pointcontrast} is too simple such that it fails to capture sufficient information from the pre-trained ShapeNet dataset. (ii) The ShapeNet dataset provides point clouds with clean spatial structures and fewer noises, which benefits the pre-trained model to learn effective representations. (iii) Although the amount of sequence data in ScanNet is large, the modality might still be limited as it only has 707 scenes. This last hypothesis is further backed by our experiments in \textit{Data Efficiency} below.

\paragraph{Temporal Transformation}

As described in \cref{sec:sequence,sec:dataset}, we learn from synthetic view transformation on object shapes and natural view transformation on physical scenes. To study their effects, we disentangle the combinations by removing certain transformations to generate training data of synthetic shapes when pre-training on the ShapeNet dataset; \cref{tab:ablation-view-syn} summarizes results. For physical scenes, we pre-train PV-RCNN on the KITTI dataset and compare the models trained with and without sampling input data from natural sequences; \cref{tab:ablation-view-natural} summarizes the results. Temporal transformation introduces substantial performance gains in both cases.

\begin{table}[htb!]
    \centering
    \caption{\textbf{Ablation study: temporal transformation}}
    \label{tab:ablation-view}
        \begin{subtable}[h]{\linewidth}
            \centering
            \caption{\textbf{Synthetic Shapes.} We evaluate the pre-trained PointNet model on ModelNet40 by linear evaluation under different temporal transformations.}
            \label{tab:ablation-view-syn}
            \scalebox{0.85}{%
                \begin{tabular}{cccc}
                    \toprule
                    Synthetic View Transformations & Accuracy\\
                    \midrule
                    Full & \textbf{88.3\%} \\
                    Remove rotation & 87.8\%\\
                    Remove scaling & 87.9\% \\
                    Remove translation & 87.2\% \\
                    Remove rot. + sca. + trans.  & 85.5\% \\
                    \bottomrule 
                \end{tabular}%
            }%
    \end{subtable}
    \begin{subtable}[h]{\linewidth}
        \centering
        \caption{\textbf{Physical Scenes.} We freeze the PV-RCNN backbone and fine-tune the 3D object detector on KITTI. It shows the mAP results (under 40 recall positions) of car detection w./w.o. sampling input data from natural sequence.}
        \label{tab:ablation-view-natural}
        \scalebox{0.85}{%
            \begin{tabular}{cccc}
                \toprule
                \multirow{2}{*}{Natural Sequence} & \multicolumn{3}{c}{Car}\\
                 & Easy & Moderate & Hard \\
                \midrule
                \ding{51} & \textbf{91.08} & \textbf{81.63} & \textbf{79.39}\\
                \ding{55} & 90.17 & 81.21 & 79.05 \\
                \bottomrule 
            \end{tabular}%
        }
    \end{subtable}
\end{table}

\paragraph{Spatial Data Augmentation}

We investigate the effects of spatial data augmentations by turning off certain types of augmentation; see \cref{tab:ablation-spatial}. By augmenting the point clouds into different shapes and dimensions, random crop boosts the performance, whereas random cutout hurts the performance as it breaks the point cloud's structural continuity, crucial for point-wise feature aggregation from neighbors.

\begin{table}[htb!]
    \centering
    \caption{\textbf{Ablation study: spatial data augmentation.} We pre-train the PointNet model on ShapeNet with different spatial transformations. Performances below reflect the linear evaluation results on ModelNet40.}
    \label{tab:ablation-spatial}
    \scalebox{0.85}{%
        \begin{tabular}{cc}
            \toprule
             Spatial Transformation & Accuracy\\
            \midrule
             Full & \textbf{88.3}\%\\
             Remove Cutout & 88.1\% \\
             Remove Crop & 87.5\%\\
             Remove Crop and Cutout & 87.4\%\\
             Down-sample only & 86.1\%\\
            \bottomrule
        \end{tabular}%
    }%
\end{table}

\paragraph{Data Efficiency}

To further analyze how the size of training data affects our model, we pre-train the DGCNN model with a subset of ScanNet dataset by sampling 25,000 frames depth images from the entire 1,513 sequences. Evaluated on ModelNet40, the model's performance only drops about $0.5$\% for both linear evaluation and fine-tuning compared with training on the whole set with 0.4 million frames; such results are similar to 2D image pre-training~\cite{he2020momentum}. We hypothesize that increasing the data diversity, but not sampling density, would improve performances in self-supervised 3D representation learning. 

\paragraph{Robustness}

We observe that the proposed \ac{strl} can learn the self-supervised representations by simple augmentations; it robustly achieves a satisfying accuracy (about 85\%) on ModelNet40 linear classification. Nevertheless, it differs from the results shown in 2D image pre-training~\cite{chen2020simple,grill2020bootstrap}, where data augmentations affect the ImageNet linear evaluation by up to 10\%. We hypothesize that this difference might be ascribed to the general down-sampling process performed on the point clouds, which introduces structural noises and helps the invariant feature learning.

\section{Conclusion}

In this paper, we devise a spatio-temporal self-supervised learning framework for learning 3D point cloud representations. Our method has a simple structure and demonstrates promising results on transferring the learned representations to various downstream 3D scene understanding tasks. In the future, we hope to explore how to extend current methods to holistic 3D scene understanding~\cite{huang2018holistic,huang2018cooperative,huang2019perspectivenet,chen2019holistic++,jia2020lemma,qi2018human,jiang2018configurable} and how to bridge the domain gap by joint training of unlabeled data from various domains.

{
\setstretch{1}
\small
\balance
\bibliographystyle{ieee_fullname}
\bibliography{reference}
}

\end{document}

%% file: config.tex
\usepackage{times}
\usepackage{graphicx}
\usepackage{amsmath,amssymb,amsthm,mathabx}
\usepackage{booktabs}
\usepackage{algorithmic}
\usepackage[linesnumbered,ruled,vlined]{algorithm2e}
\usepackage{acronym}
\usepackage{enumitem}
\usepackage[pagebackref=true,breaklinks=true,colorlinks,bookmarks=false]{hyperref}
\usepackage{balance}
\usepackage{pifont}
\usepackage{setspace}
\usepackage[skip=3pt,font=small]{subcaption}
\usepackage[dvipsnames]{xcolor}
\usepackage[capitalise]{cleveref}
\usepackage{tabularx,multirow,array,makecell}
\usepackage{overpic,wrapfig}

\makeatletter
\renewcommand{\paragraph}{%
  \@startsection{paragraph}{4}%
  {\z@}{0ex \@plus 0ex \@minus 0ex}{-1em}%
  {\hskip\parindent\normalfont\normalsize\bfseries}%
}
\makeatother

\graphicspath{{figures/}}

\crefname{algocf}{Alg.}{Algs.}
\Crefname{algocf}{Algorithm}{Algorithm}
\crefname{section}{Sect.}{Sect.}
\Crefname{section}{Sect.}{Sect.}
\crefname{subsection}{Sect.}{Sect.}
\Crefname{subsection}{Sect.}{Sect.}
\crefname{subsubsection}{Sect.}{Sect.}
\Crefname{subsubsection}{Sect.}{Sect.}

\definecolor{gblue}{HTML}{4285F4}
\definecolor{gred}{HTML}{DB4437}

\acrodef{strl}[STRL]{\underline{s}patio-\underline{t}emporal \underline{r}epresentation \underline{l}earning}

\SetKwInput{KwInput}{Input}                
\SetKwInput{KwOutput}{Output}              

\SetCommentSty{mycommfont}

\newcommand\blfootnote[1]{%
  \begingroup
  \renewcommand\thefootnote{}\footnote{#1}%
  \addtocounter{footnote}{-1}%
  \endgroup
}